\documentclass[sigconf]{acmart}

\usepackage{bm}
\usepackage{subfigure}

\newcommand{\argmax}{\mathop{\rm arg~max}\limits}

\iftrue 

\else

\fi

\copyrightyear{2024}
\acmYear{2024}
\setcopyright{acmlicensed}\acmConference[MOCO '24]{9th International Conference 
on Movement and Computing}{May 30-June 2, 2024}{Utrecht, Netherlands}
\acmBooktitle{9th International Conference on Movement and Computing (MOCO '24), 
May 30-June 2, 2024, Utrecht, Netherlands}
\acmDOI{10.1145/3658852.3659076}
\acmISBN{979-8-4007-0994-4/24/05}

\acmConference[MOCO '24]{The 9th International Conference on Movement and Computing}{May 30--June 2, 2024}{Netherlands}
\acmPrice{15.00}

\begin{document}

\title{Automatic Dance Video Segmentation \\for Understanding Choreography}

\author{Koki Endo}
\authornote{These authors contributed equally and are ordered alphabetically}
\affiliation{%
  \institution{The University of Tokyo}
  \city{Tokyo}
  \country{Japan}
}
\email{endo-koki@is.s.u-tokyo.ac.jp}

\author{Shuhei Tsuchida}
\authornotemark[1]
\affiliation{%
  \institution{Ochanomizu University}
  \city{Tokyo}
  \country{Japan}
}
\email{tsuchida.shuhei@ocha.ac.jp}

\author{Tsukasa Fukusato}
\affiliation{%
  \institution{Waseda University}
  \city{Tokyo}
  \country{Japan}
}
\email{tsukasafukusato@waseda.jp}

\author{Takeo Igarashi}
\affiliation{%
  \institution{The University of Tokyo}
  \city{Tokyo}
  \country{Japan}
}
\email{takeo@acm.org}


\begin{abstract}
Segmenting dance video into short movements is a popular way to easily understand dance choreography.
However, it is currently done manually and requires a significant amount of effort by experts.
That is, even if many dance videos are available on social media (e.g., TikTok and YouTube), it remains difficult for people, especially novices, to casually watch short video segments to practice dance choreography.
In this paper, we propose a method to automatically segment a dance video into each movement.
Given a dance video as input, we first extract visual and audio features: the former is computed from the keypoints of the dancer in the video, and the latter is computed from the Mel spectrogram of the music in the video.
Next, these features are passed to a Temporal Convolutional Network (TCN), and segmentation points are estimated by picking peaks of the network output.
To build our training dataset, we annotate segmentation points to dance videos in the AIST Dance Video Database, which is a shared database containing original street dance videos with copyright-cleared dance music.
The evaluation study shows that the proposed method (i.e., combining the visual and audio features) can estimate segmentation points with high accuracy.
In addition, we developed an application to help dancers practice choreography using the proposed method.
\end{abstract}





\maketitle


\section{Introduction}
There are two ways for people to practice dance choreography: one is to be taught by an instructor, and the other is to imitate the choreography in dance videos.
In instructor-based practice, the instructor often divides the choreography into several segments, taking into account the structure of the choreography, and teaches them one by one for ease of practice.
This segmentation idea is suitable for practicing a series of movements called ``dance moves'' or movements that are smoothly connected without breaks as one segment, but people must find a good instructor who segments the dance choreography that they want to practice and teaches each segment.
On the other hand, in the case of video-based practice, people can practice the choreography anytime and anywhere as long as they have the video.
To understand the choreography in the video, it is very important to segment the dance videos instead of the instructors.
However, manual segmentation is too difficult for novices because it requires various knowledge of dance movements.
Note that, even for expert dancers, this is time-consuming and labor-intensive while imagining the structure of the choreography in a dance video.
As a result, even if people find the desired choreography on social media (e.g., TikTok or YouTube), they often give up on the dance practice itself because there is no instructor to teach them and standard dance videos do not have any segmentation points.


Therefore, we propose a method to automatically segment a dance video into short movements.
Since dance choreography is related to music, we assume that both visual and audio information can be used for the segmentation task and extract visual and audio features from an input dance video.
The visual feature is obtained by calculating bone vectors from the keypoints of the dancer and then passing the vectors to a fully connected network, and the audio feature is computed by passing the Mel spectrogram of the music in the video to a two-dimensional CNN.
Next, we compute a segmentation probability score in each video frame from these features based on a Temporal Convolutional Network (TCN)~\cite{Bai2018TCN}.
Finally, the segmentation points are estimated by picking peaks of the probability scores.
Our method does not require any genre-specific knowledge and thus can be applied to any genre of dance.

For the training and evaluation of the proposed model, we first developed a simple tool to annotate segmentation points for the videos.
Based on the tool, we constructed a dataset by manually annotating segmentation points for dance videos in the AIST Dance Video Database~\cite{Tsuchida2019AISTDanceDB} by the first author of this paper (who has prior experience in dance) and 20 other dancers.
Considering the ambiguity of the segmentation of each annotator, we defined the ground truth of the segmentation points, named segmentation probability, by representing each segmentation as the sum of Gaussian distributions.
We evaluated the proposed method using the dataset and found that the proposed method can correctly estimate the segmentation points.
We also confirmed the effectiveness of the visual and audio features through an ablation study.

Furthermore, we developed an application that helps users practice dance choreography using dance videos.
The application automatically segments dance videos using the proposed method and can play back each segment, allowing users to efficiently practice dance choreography segment by segment. 
In addition, the application enables users to intuitively explore segmentation timings according to their preferences.

Our principal contributions are summarized as follows:
\begin{enumerate}
    \item We propose a method to automatically segment dance videos into short segments by using both visual and audio features. The evaluation results show the effectiveness of both features for the segmentation task.
    \item We collect segmentation candidates from 21 dancers and construct a dataset with 1410 pairs of dance videos and corresponding segmentation point labels.
    \item We propose an application to help users practice dance with a dance video using our segmentation method.
\end{enumerate}

\begin{figure*}[t]
  \includegraphics[width=0.8\textwidth]{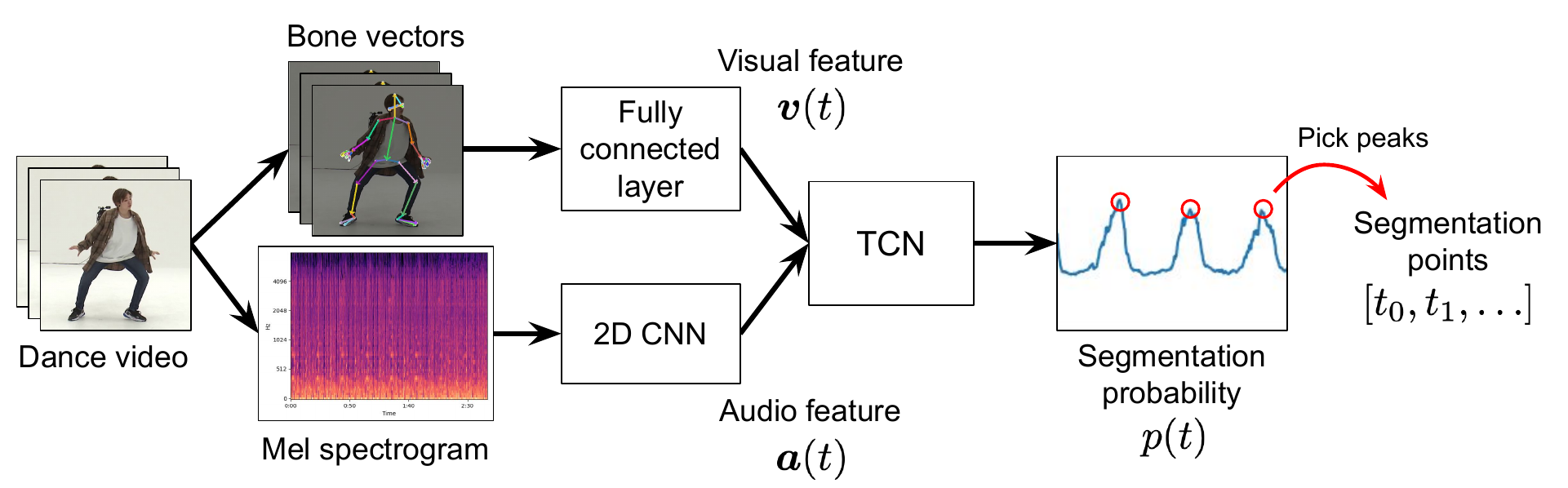}
  \caption{Overview of the proposed method. First, we extract visual and audio features from an input dance video. Then, the system automatically estimates the segmentation probability score based on a Temporal Convolutional Network (TCN), and final segmentation points can be obtained by simply picking peaks of the probability.}
  \Description{Diagram showing the overview of the proposed algorithm. As the start node, a stack of three dancer images labeled “Dance video” is on the left. Two arrows extend from this node, one to a stack of three dancer images labeled “Bone vectors” and the other to an image of a Mel spectrogram labeled “Mel spectrogram.” The “Bone vectors” feed into a node labeled “Fully connected layer,” and the “Mel spectrogram” feeds into a node labeled “2D CNN.” The “Fully connected layer” and “2D CNN” are connected to a node labeled “TCN” by arrows labeled “Visual feature v(t)” and “Audio feature a(t),” respectively. The “TCN” feeds into an image of a line graph labeled “Segmentation probability p(t).” In this image, the graph has three peaks circled in red. The “Segmentation probability p(t)” is connected to the end node “Segmentation points [t0, t1, …]” by an arrow labeled “Pick peaks.”}
  \label{fig:algo}
\end{figure*}

\section{Related Work}

\subsection{Effectiveness of Dance Segmentation}
Rivière et al.~\cite{Riviere2019Decomposition} observed how dancers practiced choreography with a dance video.
This observation showed that the dancers often segmented the choreography into short movements to practice.
Moreover, they compared the segmentation by the dancer, who practices the choreography, with that by the teacher, who teaches the choreography to the dancer.
As a result, for novice dancers, the segmentation by the teacher was more effective in learning the choreography.
These results suggest that properly segmented dance videos help novice dancers practice the choreography.
Therefore, we aim to automate this segmentation task.

\subsection{Dance Motion Segmentation}
There is some research related to automatic dance motion segmentation.
Shiratori et al.~\cite{Shiratori2004DanceMotionStructure} proposed a method to segment traditional Japanese dance motion from motion data and musical information based on some heuristic rules.
However, this method cannot apply to other kinds of dance because these heuristic rules are specific to traditional Japanese dance.
Okada et al.~\cite{Okada2015DanceSegmentation} segmented dance motion into short movements for reusing in the generation of 3DCG character animation.
Their method also employs a limited rule-based approach.
In addition, the method requires known musical beats for the segmentation, so it is difficult to apply to arbitrary dance videos, where the musical tempo is unknown.

Therefore, we extract visual and audio features from an input dance video and automatically segment it using a neural network (without specific dance knowledge).

\subsection{Support for Dance Practice}
When practicing dance choreography, dancers first try to understand the choreography and then refine their movements.
There are various research and tools to support each of these two phases.

For understanding choreography, some research helps the practice by editing dance videos or devising a novel learning method.
Saito et al.~\cite{Saito2020Onomatopee} manually annotated onomatopoetic words to dance videos so that dancers can easily understand the nuance of dance movements.
Fujimoto et al.~\cite{Fujimoto2012DanceSelfImage} and Tsuchida et al.~\cite{Tsuchida2022DanceSelfModeling} generated a video that looks as if the users themselves have mastered the choreography in the target video.
Their research is based on the hypothesis that the generated video is easier for the users to understand and practice the choreography than the original target video.
Tsuchida et al.~\cite{Tsuchida2022SeparatedLearning} proposed a method to separate the dance practice process into two phases: dancers first receive information on the movement they try to practice without moving their body and then move their body.
This idea is simple but enables users to save the working memory in their own brains.

Some applications and tools also help dancers understand choreography.
Ugotoru~\cite{Ugotoru} can flip a dance video horizontally and change the playback speed, but it cannot segment the video.
SymPlayer~\cite{SymPlayer} can play a part of the video on a loop by manually specifying the start and end of the part.
MoveOn~\cite{Riviere2019Decomposition} enables users to manually segment a video and play each segment.
However, since the manual segmentation requires various knowledge of the choreography, it remains difficult for users, even experts, to use this tool.

We therefore propose a method to automatically segment dance videos, enabling people to easily practice the choreography.

\subsection{Temporal Convolutional Network}
Temporal Convolutional Networks (TCNs)~\cite{Bai2018TCN} are network architectures that convolve sequential data with one-dimensional kernels along the time axis.
The main characteristic of TCNs is a hierarchical dilated convolution: they can have a large receptive field by increasing the interval between the elements to be convolved as the layer gets deeper.

TCNs can be causal or non-causal.
The former convolves only past and current elements and is suitable for real-time processing.
On the other hand, the latter convolves past, current, and future elements and can be used for offline computation.
Our method adopted a non-causal TCN to improve the model performance using future information because we do not need real-time computation.

Each layer of a TCN has a structure of a residual network~\cite{He2016Residual}, which outputs the addition of the transformed result and the original input.
He et al.~\cite{He2016Residual} and Bai et al.~\cite{Bai2018TCN} showed that the residual network converged faster and performed better than the network without the residual function.
Each TCN layer consists of two one-dimensional convolutions and an addition function.
In order to make convergence faster, each convolution is followed by a weight normalization~\cite{Salimans2016WeightNorm}.
In our case, the input and the convolved result can be simply added because both of them have the same size.
If the size of the result is smaller than the input, the input needs to be downsampled.

Compared with recurrent networks, TCNs can be trained faster because they have a simpler architecture and fewer parameters, and the convolution can be parallelized.
Taking advantage of this, TCNs are used in various sequence modeling.
Bai et al.~\cite{Bai2018TCN} showed that a causal TCN performed better than recurrent networks in various tasks, including music harmonization modeling and language modeling.
Davies et al.~\cite{Davies2019TcnBeatTracking} used a non-causal TCN for beat tracking, and Pedersoli and Goto~\cite{Pedersoli2020DanceBeatTracking} estimated musical beats from visual information only in the dance video with a non-causal TCN.

we then take a non-casual TCN to estimate segmentation points from dance videos.

\section{Proposed Method}

\subsection{Algorithm Overview}
Our method takes a dance video as input and estimates its segmentation points using a supervised learning-based approach.
\autoref{fig:algo} shows the overview of the proposed method.
We first extract visual and audio features from an input dance video.
The former is computed by passing the bone vectors of the dancer in the video to a fully connected layer.
The latter is computed by convolving the Mel spectrogram of the music in the video.
These features are then passed to a TCN, and the network outputs segmentation probability.
Finally, segmentation points are computed by picking peaks of the probability.
The details of training data are described in Section~\ref{sec:data}.

\subsection{Visual Feature}
A visual feature represents the dancer's movements in the video.
First, we detect body keypoints of the dancer using AlphaPose~\cite{Fang2022AlphaPose}.
AlphaPose computes pixel coordinates of the keypoints and their confidence for each video frame.
The number of the keypoints used in our method is 68 (= 26 for the whole body and 21 for each hand).

We assume that only one dancer is in the video, but AlphaPose sometimes accidentally detects multiple or no people.
In the former case, we adopt one dancer whose sum of confidence score is the greatest.
The adopted dancer index at the $t$-th frame $j^\ast(t)$ is calculated as follows:
\begin{equation}
    j^\ast(t) = \argmax_j \sum_{i=0}^{67} \gamma_{ji}(t)
\end{equation}
where $\gamma_{ji}(t)$ represents the confidence of the $i$-th keypoint of the $j$-th dancer at the $t$-th frame.

In the latter case, we linearly interpolate the keypoint coordinates.
When the $i$-th keypoint at the $t$-th frame $\bm{x}_i(t)$ is not detected, it is calculated as follows:
\begin{equation}
    \bm{x}_i(t) = \frac{t_n - t}{t_n - t_p} \bm{x}_i(t_p) + \frac{t - t_p}{t_n - t_p} \bm{x}_i(t_n)
\end{equation}
where $t_p$ is the greatest frame less than $t$ of the frames at which the keypoint is detected.
Similarly, $t_n$ is the least frame greater than $t$ of the frames at which the keypoint is detected.

Next, we compute 67 bone vectors from the keypoint coordinates.
These vectors are then normalized so that the length will be 0.5.
Thus, by converting the keypoint coordinates to normalized bone vectors, it is possible to reduce the influence of differences in dancers' body size and the camera distance/angle.


Finally, we pass the normalized vectors to a fully connected layer and an activation function to extract the relation between vectors and important information.
This layer is applied independently for each video frame.
Since the number of bone vectors is 67, and each vector has horizontal and vertical elements, the number of input channels of the layer is $134 (= 67 \times 2)$.
Then the visual feature at the $t$-th frame $\bm{v}(t) \in \mathbb{R}^{67}$ is obtained as the layer output.
The number of output channels of the layer is set to 67, which is the same as the number of vectors.
We used an Exponential Linear Unit (ELU)~\cite{Clevert2016ELU} for the activation function.

\subsection{Audio Feature}
An audio feature represents the musical information in the dance video.
Since many dance movements correspond to music, it is thought that not only visual but also musical information is important for dance segmentation tasks.
First, we compute a Mel spectrogram by applying a short-time Fourier transform (STFT) to the music in the video.
The Mel spectrogram is a two-dimensional array that represents the strength of elements for each frequency and time, and has a value from -80\,dB to 0\,dB.
We then normalize the spectrogram by linear transform so that the value will be in $[-0.5, 0.5]$.
The parameters for computing the Mel spectrogram are set to the same values as in \cite{Davies2019TcnBeatTracking}.

Next, the audio feature is obtained by passing the normalized spectrogram to a two-dimensional convolutional block.
The purpose of this convolution is to compress the spectrogram so that its time resolution will be the same as that of the video frames.
With $T$ denoting the number of video frames, and $S \in \mathbb{R}^{T \times 81}$ denoting the normalized spectrogram, the audio feature at the $t$-th frame $\bm{a}(t) \in \mathbb{R}^{16}$ is computed as follows:
\begin{equation}
    \bm{a}(t) = \mathrm{Conv}(S_{i-2: i+2})
\end{equation}
where $i$ is the index of the spectrogram sample nearest in time to $t$, $S_{i-2: i+2} \in \mathbb{R}^{5 \times 81}$ is the slice of $S$ from $i-2$ to $i+2$, and $\mathrm{Conv}$ is the convolutional block.
This block is the same as the convolutional block described in \cite{Davies2019TcnBeatTracking}, and the number of output channels is 16.

\subsection{Temporal Convolutional Network}
\begin{figure}[tbp]
    \centering
    \includegraphics[width=\hsize]{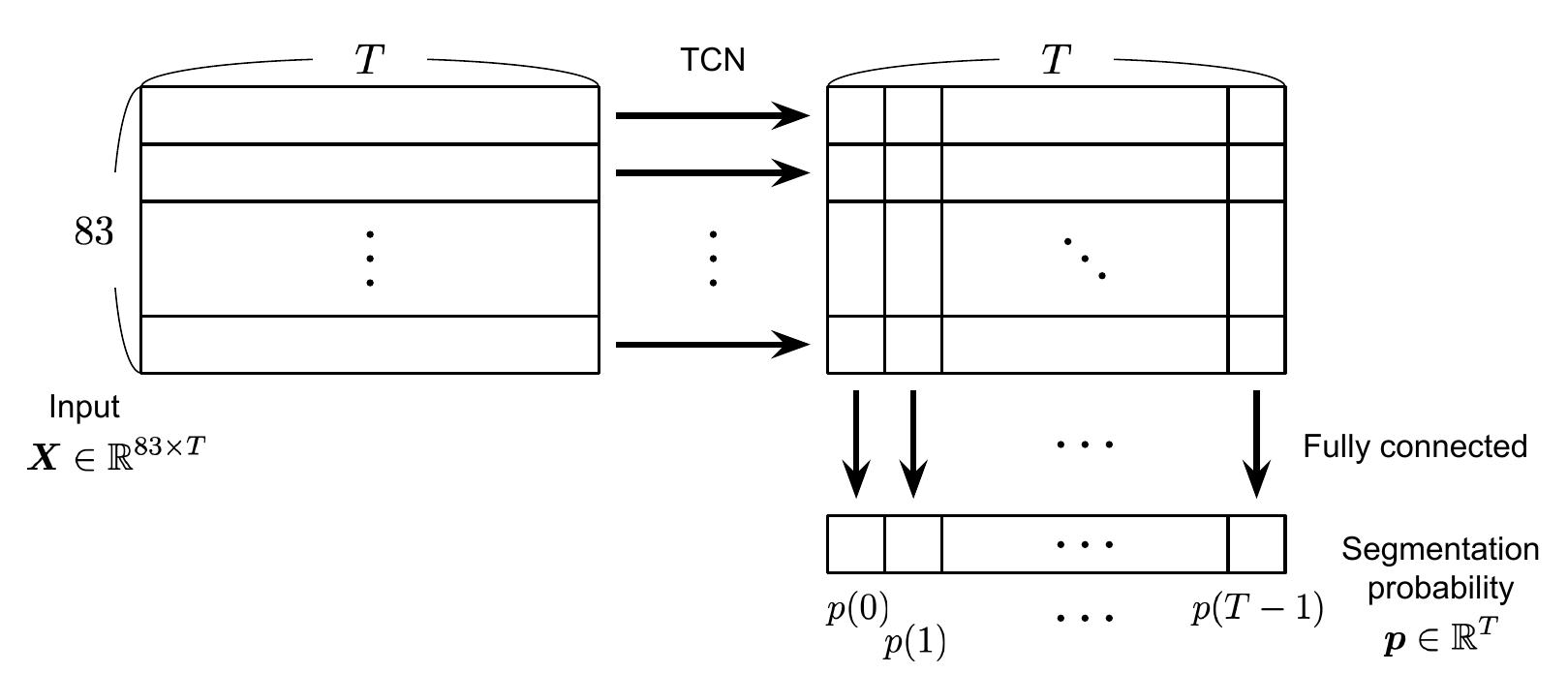}
    \caption{Overview of the proposed network. The input $\bm{X}$ is passed to the TCN for each row and then the fully connected layer for each column.}
    \Description{Diagram showing the structure of the proposed network. As input, the first rectangle representing an 83 by T matrix is on the upper left. This rectangle has horizontal ruled lines to indicate that the matrix has several rows. Each row of the first rectangle is connected to each row of the second rectangle on the upper right by right arrows labeled “TCN.” The second rectangle has the same size as the first and has horizontal and vertical ruled lines. Each column of the second rectangle is connected to each column of the third rectangle labeled “Segmentation probability” on the lower right by down arrows labeled “Fully connected.” The size of the third rectangle is the same as the single rows of the first and the second rectangles. The third rectangle has vertical ruled lines, and each column is labeled “p(0),” “p(1),” …, and “p(T-1).”}
    \label{fig:ourTCN}
\end{figure}

From the visual and audio features, we compute a segmentation probability score using a non-causal TCN (see \autoref{fig:ourTCN}).
The score means the probability that the input video could be segmented at each frame.
The input for the TCN $\bm{X}$ is represented as follows:
\begin{equation}
    \bm{X} = \begin{bmatrix}
        \bm{v}(0) & \ldots & \bm{v}(T-1) \\
        \bm{a}(0) & \ldots & \bm{a}(T-1) \\
    \end{bmatrix} \in \mathbb{R}^{83 \times T}.
\end{equation}
Because the TCN computation is applied to each input row independently, the output of the TCN is also an $83 \times T$ matrix.
Then each column of the output is passed to a fully-connected layer to obtain the segmentation probability scores $\bm{p} = \left(p(0), \ldots, p(T-1)\right) \in \mathbb{R}^T$.
The hyperparameters for the TCN are shown in \autoref{tab:tcnParam}.

\begin{table}[tbp]
    \centering
    \caption{Hyperparameters for the TCN.}
    \begin{tabular}{l|c}
        \toprule
        Number of layers & 9 \\
        Dilation width at the $i$-th layer & $2^i\ (i = 0, \ldots, 8)$ \\
        Kernel size & 5 \\
        Dropout rate & 0.1 \\
        Activation function & ELU \\
        \bottomrule
    \end{tabular}
    \Description{Table showing the hyperparameters for the TCN. The number of layers is 9. For each i = 0, …, 8, the dilation width at the i-th layer is the i-th power of 2. The kernel size is 5, and the dropout rate is 0.1. The activation function is ELU.}
    \label{tab:tcnParam}
\end{table}

\subsection{Peak Picking} \label{sec:peak}
Based on the estimated probability scores, we automatically detect segmentation points.
In this paper, we simply defined the $t^\ast$-th frame as a segmentation point if it satisfies the following two conditions: (1)~the probability score~$p(t^\ast)$ is greater than a certain threshold $h$ and (2)~the probability score~$p(t^\ast)$ is the greatest in a local window centered at $t^\ast$.
These conditions are represented as follows:
\begin{equation}
    \Bigl\{p(t^\ast) > h \Bigr\} \wedge \Bigl\{p(t^\ast) = \max_{t^\ast - w/2 \le t \le t^\ast + w/2} p(t) \Bigr\}
\end{equation}
where $w$ represents a local window size (In this paper, we empirically set $w = 20$).

\section{Dataset} \label{sec:data}

Training the proposed network model requires a dataset that contains a lot of input dance videos and the segmentation points for the videos.
We focused on the AIST Dance Video Database~\cite{Tsuchida2019AISTDanceDB} and constructed a new dataset by manually annotating segmentation points into its videos.
The original AIST Dance Video Database contains various genres of dance (i.e., break, house, ballet jazz, street jazz, krump, LA-style hip-hop, lock, middle hip-hop, pop, and waack), and its videos are categorized into 7 types (i.e., basic dance, advanced dance, group dance, moving camera, showcase, cypher, and battle).
In this paper, we selected two types of videos, ``basic'' and ``advanced.'' 
In basic dance videos, a dancer performs the basic movement specific to each genre.
Each dance is 16 beats long, and the average video length is 23 seconds.
In advanced dance videos, a dancer performs their original choreography that contains various movements.
Each dance is 64 beats long, and the average video length is 52 seconds.
The total number of dance videos in our dataset is 1410 (1200 for the basic dance and 210 for the advanced dance).
Note that the resolution of each video is $1920 \times 1080$ pixels and the frame rate is $59.94$ [fps].

\subsection{Annotation Tool}
\subsubsection{User Interface} \label{sec:annotationUI}
\begin{figure}[tbp]
    \centering
    \includegraphics[width=\hsize]{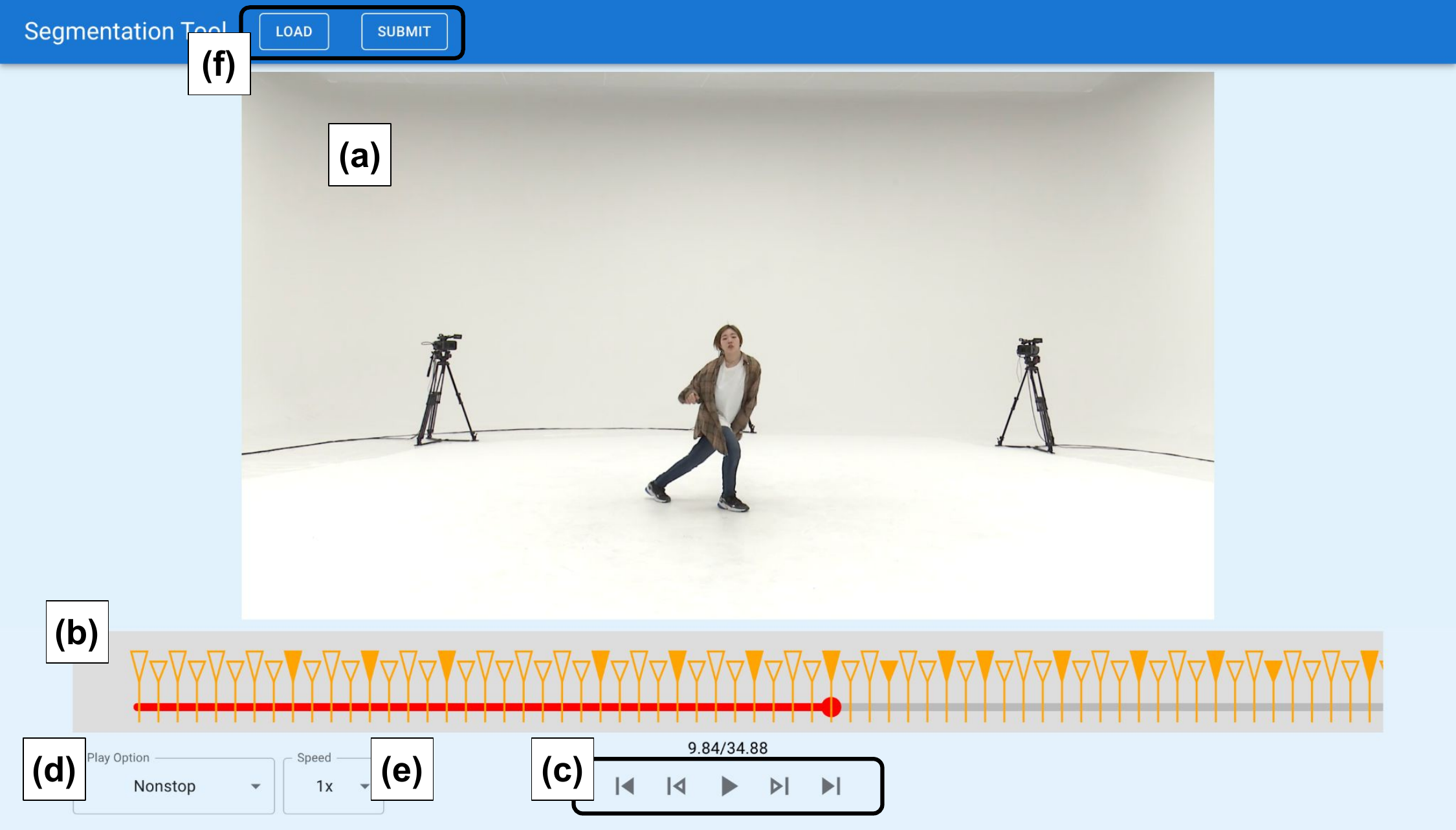}
    \caption{User interface of our annotation tool. (a) Dance video. (b) Seek bar and segmentation candidates. (c) Playback and skip buttons. (d) Pull-down to change the playback mode. (e) Pull-down to change the playback speed. (f) Buttons to load a video and submit an annotation.}
    \Description{User interface of our annotation tool. A dance video is in the center (a), and a seek bar with orange markers is below the video (b). Playback and skip buttons are on the lower center (c). On the lower left, there are pull-downs to change the playback mode (d) and speed (e). This interface also has a header titled “Segmentation Tool” with two buttons (f).}
    \label{fig:segTool}
\end{figure}

We developed a simple tool to manually annotate segmentation points for each video in our dataset, as a web application (see \autoref{fig:segTool}).
The tool consists of a main panel to display a dance video in the center (\autoref{fig:segTool}a) and a seek bar with segmentation candidates (\autoref{fig:segTool}b).
Each candidate has a flag indicating whether it has been ``selected'' or ``unselected.''
Note that all the flags are initially set to ``unselected.''

As with standard video players, the users can play or pause the video by clicking a playback button (see \autoref{fig:segTool}c).
The skip buttons are located on both sides of the playback button (\autoref{fig:segTool}c), and enables users to skip the video to the nearest segmentation candidate to the current playback time by clicking the unfilled skip buttons.
The filled skip buttons are similar, but skip the video to the ``selected'' candidate closest to the current playback time.
In addition, the users can change the playback mode with a pull-down on the lower left (\autoref{fig:segTool}d).
The tool has two playback modes: {\it Nonstop} and {\it Pause}.
In the {\it Nonstop} mode, the video continues to play until the end unless paused by the users.
In the {\it Pause} mode, the video pauses automatically when the playback time reaches the selected segmentation candidates, which helps the users to confirm their segmentation.
The users can also change the playback speed with a pull-down on the lower left (\autoref{fig:segTool}e).

As an annotation function, the user clicks on candidates and the system then flags them as ``selected'' and colors them orange.
In contrast, when clicking on a ``selected'' candidate again, the system assigns a transparent color to it and flags them as ``unselected.''
After setting the candidate flags, the users can export the ``selected'' candidates as segmentation points (see \autoref{fig:segTool}f).

\subsubsection{Segmentation Candidates}
It is desirable to set appropriate segmentation candidates so that the users can simply efficiently annotate segmentation points into each video.
Then, we used the timings of beats and half-beats of the music in the video as the segmentation candidates.
Since the tempo information of the music in the AIST Dance Video Database is known and the audio in the video has already been denoised, the timing of the $i$-th segmentation candidate $c_i$ is computed as follows:
\begin{equation}
    c_i = c_0 + \frac{60}{\tau} \cdot \frac{i}{2}\quad (i = 0, \ldots, 2n_\mathrm{b} - 1)
\end{equation}
where $c_0$ is the first timing when the absolute amplitude of the audio is greater than $0.5$, and $\tau$ is the tempo of the music in BPM.
$n_\mathrm{b}$ is the number of beats in the video and is $16$ for the basic dance videos and $64$ for the advanced dance videos.
As shown in \autoref{fig:segTool}b, the segmentation candidates at the timings of beats and half-beats are visualized as big markers and small markers, respectively.

\subsection{Data Annotation}
To obtain reliable annotations, we need to collect the annotation data from multiple dancers because where to segment the video may vary from person to person.
Then, the first author of this paper (P0), who has more than ten years of dance experience, spent 50 hours annotating the segmentation points for all the 1410 videos.
In addition, we conducted a data collection experiment with 20 participants (P1, P2, $\dots$, P20) aged 19-25.
All participants had 5 to 19 years of dance experience.

First, we explained to the participants the task details and how to use our annotation tool.
We also showed examples of segmentation points created by the first author to give the participants an idea of how to segment the video.
This explanation was conducted by using video conference and took an average of approximately 30 minutes.
Next, the participants were asked to perform a confirmation task.
In this task, the participants segmented the two videos: one is the basic dance\footnote{\url{https://aistdancedb.ongaaccel.jp/v1.0.0/video/10M/gBR_sBM_c09_d04_mBR0_ch01.mp4}}, and the other is the advanced dance\footnote{\url{https://aistdancedb.ongaaccel.jp/v1.0.0/video/10M/gMH_sFM_c09_d24_mMH3_ch18.mp4}}.
These videos are taken from the lower front camera and are not included in the dataset.
This task can check differences in segmentation strategy between participants.

After that, the participants were asked to annotate segmentation points for the videos in the dataset.
We assigned one of the ten genres to each participant, and the participants segmented all 141 videos in the assigned genre.
Since there are ten genres and 20 participants, each genre was assigned to two participants.
Therefore, at the end of the experiment, there are three annotations for each video (by the first author and two participants).
Each participant did the confirmation and the annotation tasks with his/her PC on their own time.

\subsection{Result}

\subsubsection{Difference in Confirmation Task} \label{sec:confirm}
\begin{figure}[tbp]
    \centering
    \includegraphics[width=\hsize]{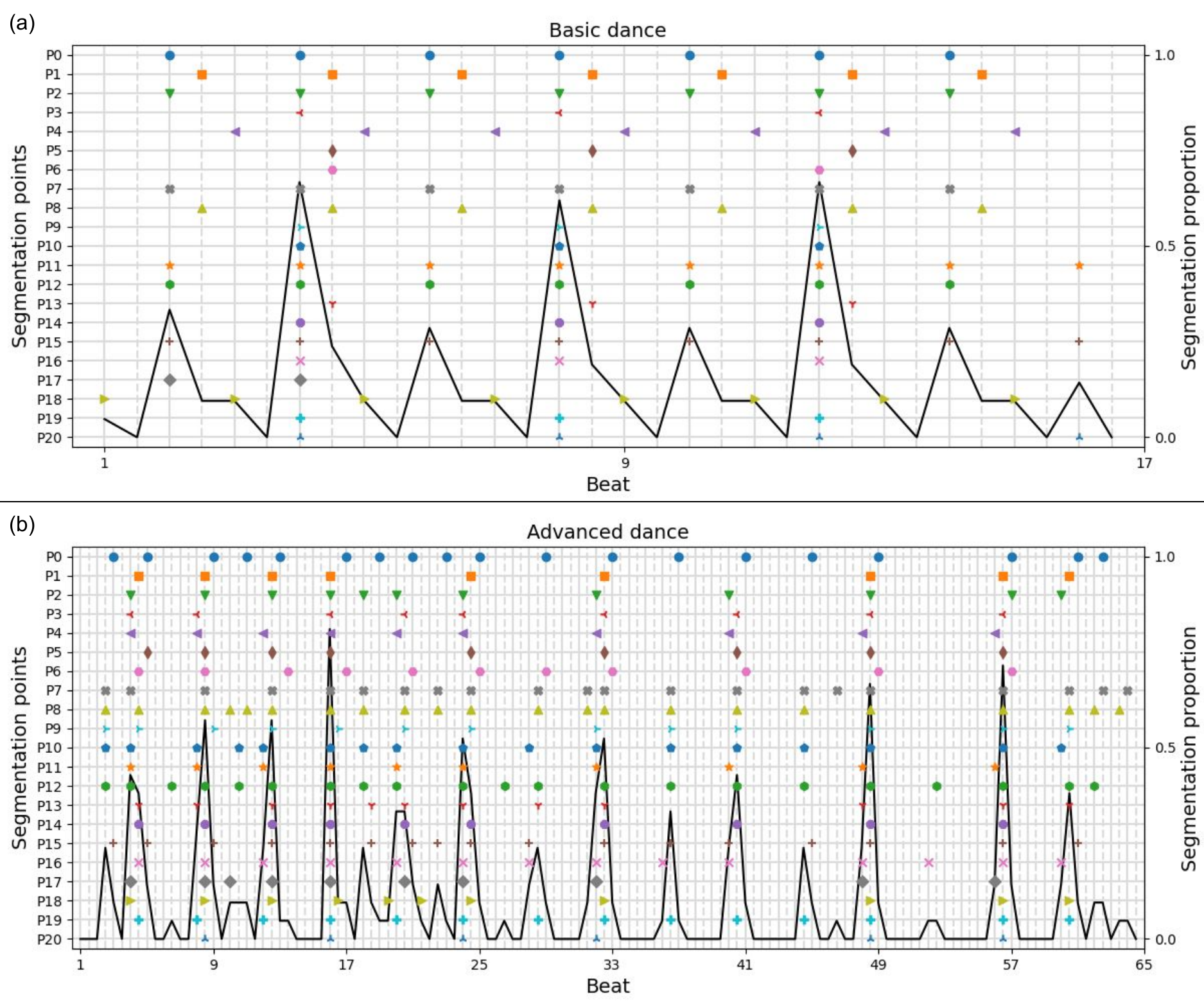}
    \caption{Segmentation points annotated by each participant. (a)~Basic dance. (b)~Advanced dance. Different marker shapes and colors represent different participants. A black line represents a segmentation proportion.}
    \Description{Scatter plots plotted at the locations designated as segmentation points by each participant in the basic dance (a) and the advanced dance (b). The Y-axis represents each participant (P0 to P20), and the X-axis represents the beat from 1 to 17 in (a) and 65 in (b). This figure also shows a segmentation proportion from 0.0 to 1.0 on the Y-axis against the beat on the X-axis. The proportion has high peaks at beats 4, 8, and 12 in (a), and 8.5, 12.5, 16, 24, 32.5, 48.5, and 56.5 in (b).}
    \label{fig:practiceGraph}
\end{figure}

Since the dancer's pose in the video looks the same even after a few frames, the participants (annotators) may have selected different segmentation candidates, even if their intended segments mean the same.
Then, we checked the differences between participants based on the results of the confirmation task.
\autoref{fig:practiceGraph} shows the segmentation points annotated by each participant and a segmentation proportion.
The segmentation proportion $\tilde{s}(b)$ represents the proportion of the dancers who segmented the video at the beat $b\ (b = 1, 1.5, \ldots, n_\mathrm{b})$ and is calculated as follows:
\begin{align}
    \tilde{s}(b) &= \frac{1}{n_\mathrm{p}} \sum_{i=0}^{n_\mathrm{p} - 1} s_i(b), \\
    s_i(b) &= \begin{cases}
        1 & (\text{P$i$ segmented the video at the beat $b$.}) \\
        0 & (\text{Otherwise.})
    \end{cases}
\end{align}
where $n_\mathrm{p}$ is the number of participants ($n_\mathrm{p} = 21$).

\begin{figure}[tbp]
    \centering
    \includegraphics[width=0.7\hsize]{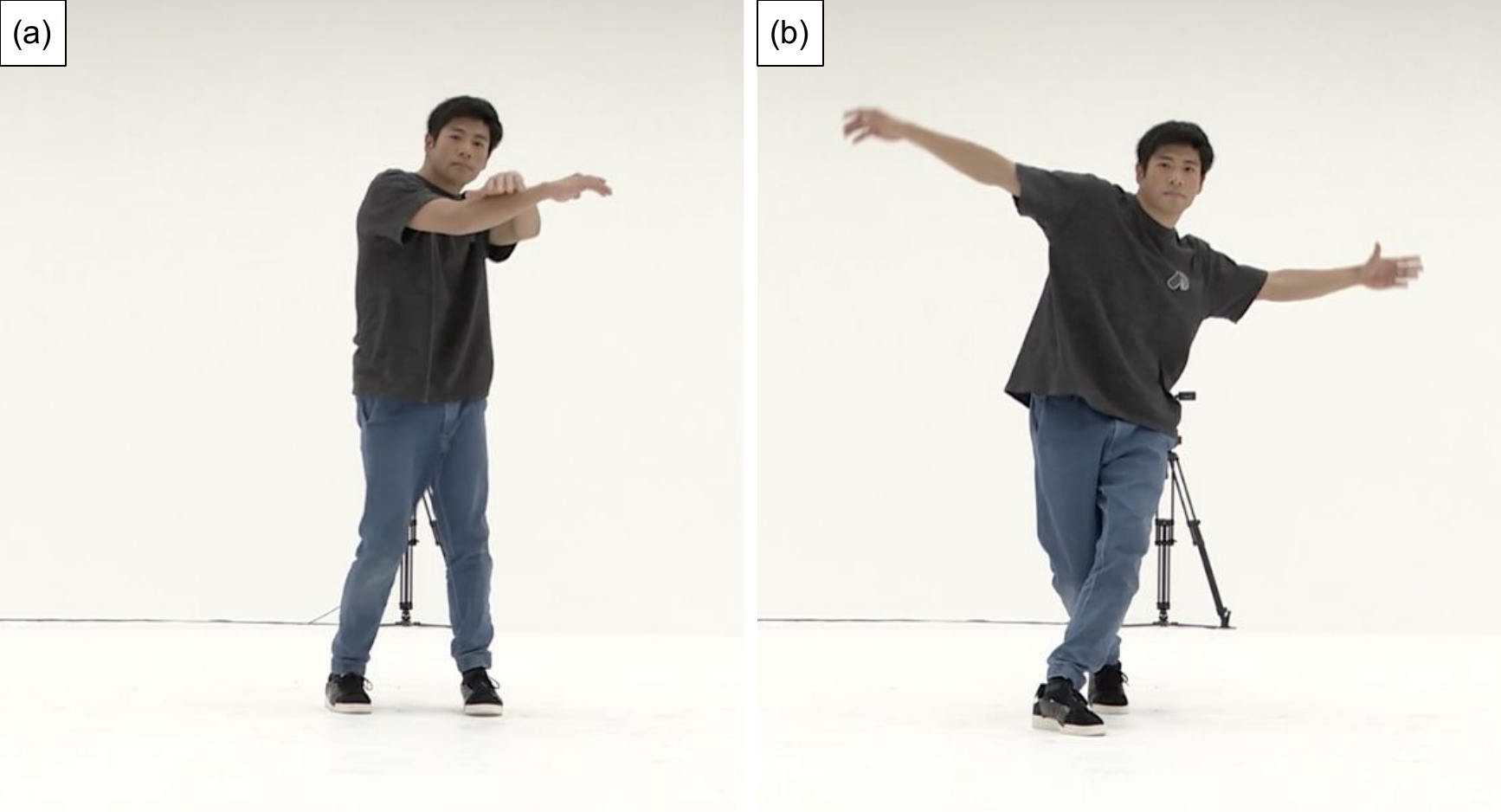}
    \caption{Video frames of the basic dance for the confirmation task. (a) At beat 12. (b) At beat 13.}
    \Description{Two video frames taken from a basic dance video. In (a), a dancer in the frame stands upright with his arms extended forward and crossed. In (b), the same dancer crosses his legs and stretches his arms to the side.}
    \label{fig:frameComparison}
\end{figure}

Since most participants set segmentation points at the same timings and the others also set segmentation points on the previous or next candidates where most participants segmented, we confirm that there are some common tendencies in dance segmentations, as shown in \autoref{fig:practiceGraph}.
For example, 14 participants segmented the basic video at beat 12 while four participants segmented at beat 12.5.
This is because, as some participants mentioned, there is ambiguity as to which candidate to select when the desired segmentation point is between the candidates.
This suggests that two segmentation points that are half-beat apart can be treated as the same segmentation.
On the other hand, two segmentation points can be distinguished if they are one beat apart.
For example, P4 and P18 segmented the basic video at beat 13, which is one beat after the candidate selected by 14 participants.
As shown in \autoref{fig:frameComparison}, the poses of the dancer in the video at beats 12 and 13 are greatly different.
Therefore, it seems reasonable to distinguish the segmentation at beats 12 and 13.
Moreover, P7 and P8 segmented the advanced video at beats 31.5 and 32.5.
This also suggests that two segmentation points that are one beat apart are distinguishable.
In summary, when a participant segmented the video at a certain point, it is assumed that the intended segmentation point is in the range of a half-beat away from that point but not one beat away.

\subsubsection{Creation of Ground Truth}
Based on Section \ref{sec:confirm}, we made a ground truth of segmentation labels for each video by representing each participant's annotations as the sum of Gaussian distributions.
This label represents the probability that the video is segmented at each frame.
First, when the $i$-th participant segmented the video at frames $[t_0, t_1, \ldots]$, a segmentation label at the $t$-th frame $l_i(t)$ is represented as follows:
\begin{align}
    l_i(t) &= \sum_j \frac{G(t \mid t_j, \sigma)}{G(0 \mid 0, \sigma)} \label{eq:l_i(t)}, \\
    G(t \mid \mu, \sigma) &= \frac{1}{\sqrt{2\pi\sigma^2}}\exp\left(-\frac{(t - \mu)^2}{2\sigma^2}\right)
\end{align}
where the denominator of the right side of \autoref{eq:l_i(t)} is to make $l_i(t)$ fall in $[0, 1]$, and $\sigma$ is set to one-third of a beat because $G(t \mid \mu, \sigma)$ is almost 0 when $| t - \mu |$ is greater than $3\sigma$.
This is based on the assumption in the previous section that the intended segmentation point is not one beat away from the selected point.
Based on $l_i(t)$, the ground truth label $\tilde{l}(t)$ can be defined as follows:
\begin{equation}
    \tilde{l}(t) = \frac{1}{n_\mathrm{label}} \sum_{i=0}^{n_\mathrm{label} - 1} l_i(t)
\end{equation}
where $n_\mathrm{label}$ is the number of segmentation labels for each video (In this paper, $n_\mathrm{label} = 3$).

\section{Evaluation}

\subsection{Experiment Setting}
We numerically evaluated the proposed method using the dataset described in Section \ref{sec:data}.
First, we divided the dataset randomly into three categories (training, validation, and test data) at a ratio of $3:1:1$.
To reduce bias due to the splitting method, the division process was done 10 times, and $10$ different divisions of data $D_i,\ i \in \{0, \ldots, 9\}$, were prepared that differ from each other in terms of which video belongs to which category.
Note that each of the three categories contains both the basic and the advanced videos at a ratio of $40:7$, and each division consists of 846 training data, 282 validation data, and 282 test data. 


Next, we evaluate the proposed model using $D_0, \ldots, D_9$.
The model was trained using the training data, and the loss was computed using the validation data at the end of each epoch.
The training was stopped when the loss did not decrease for the last ten epochs.
We adopted an L1 loss for the training of the model and Adam~\cite{Kingma2015Adam} for the optimizer.
The batch size and the learning rate were set to $1$ and $0.001$, respectively.
The model was implemented using PyTorch~\cite{Paszke2019PyTorch}, and the training was executed using a GPU on Google Colaboratory~\cite{Colab}.

\subsection{Result} \label{sec:evalResult}
The numerical performance of the proposed model was computed using the test data.
In addition to a mean test loss, we computed a mean precision, a mean recall, and an F-measure by detecting segmentation points from the ground truth label and the estimated segmentation probability.
Note that for the peak picking described in Section \ref{sec:peak}, the threshold $h$ was 0.3.
In computing the precision score, each of the estimated segmentation points was treated as true positive when there exists a ground truth point a half-beat within it.
Similarly, in computing the recall score, each of the ground truth points was treated as correctly predicted when there exists an estimated point a half-beat within it.

\begin{figure}[tbp]
    \centering
    \subfigure[A success case in the basic videos (lock dance).]{
        \includegraphics[width=\hsize]{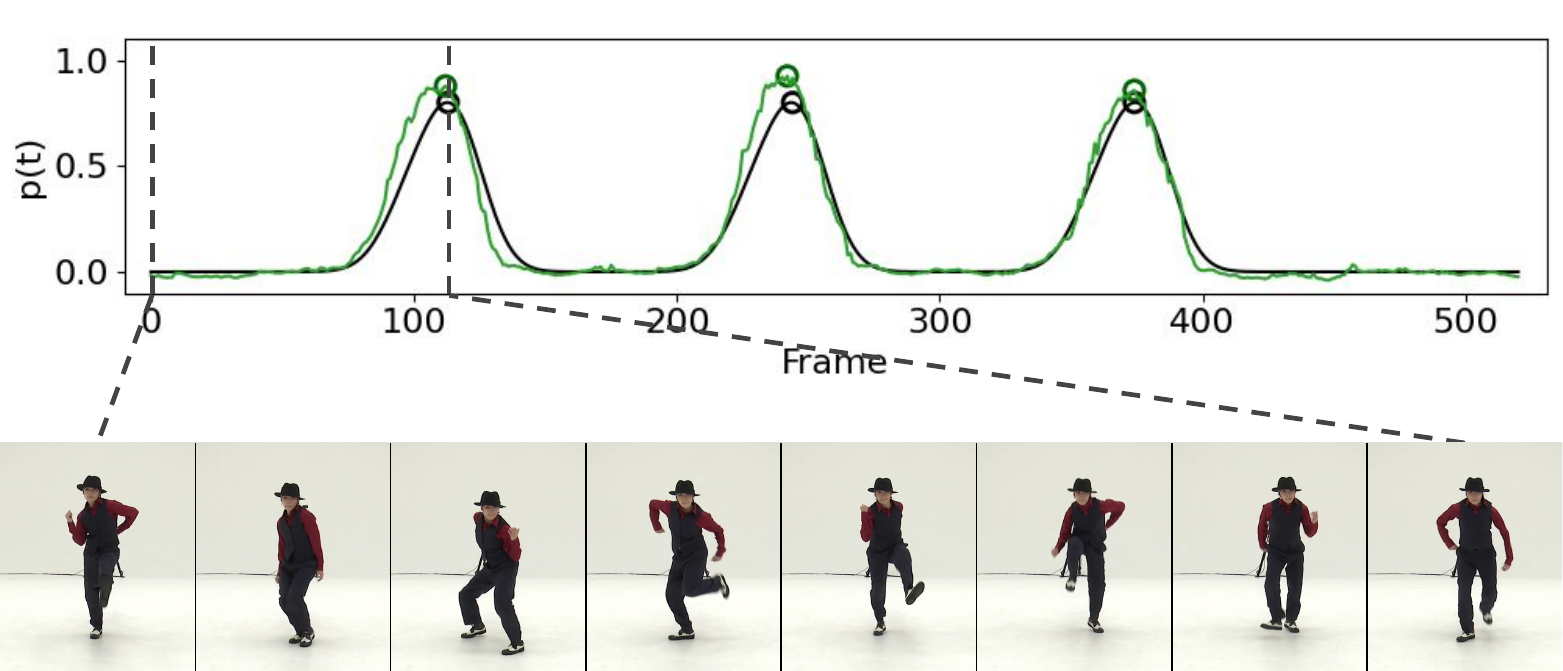}
    }
    \subfigure[A failure case in the basic videos (ballet jazz dance).]{
        \includegraphics[width=\hsize]{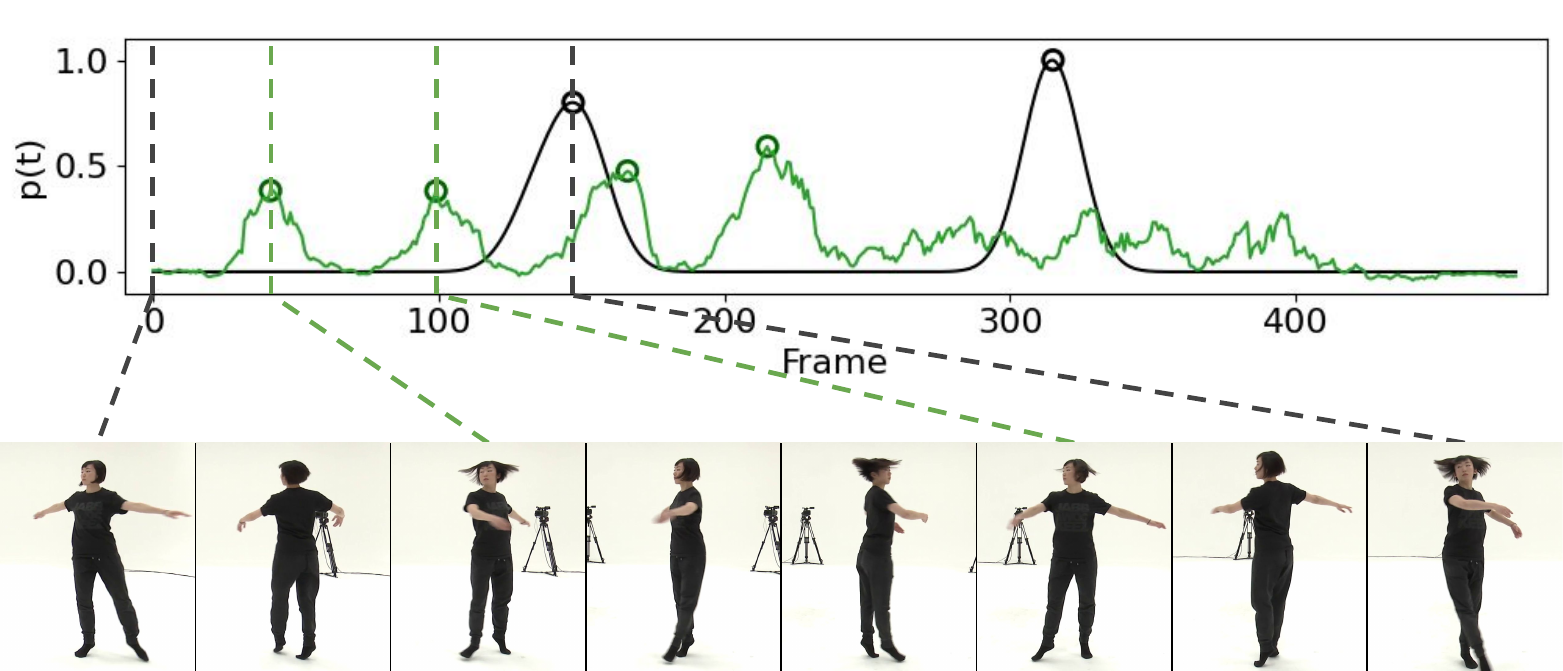}
    }
    \caption{Examples of estimating the segmentation points of the basic videos with the proposed model. Green lines are the estimated segmentation probability $p(t)$, and black lines are the ground truth label $\tilde{l}(t)$. The green and black circles are the segmentation points detected from $p(t)$ and $\tilde{l}(t)$, respectively. Snapshots are shown from the start of the video to the first segmentation point detected from $\tilde{l}(t)$.}
    \Description{Line graph showing an estimated segmentation probability and a ground truth label from 0.0 to 1.0 on the Y-axis against video frames from 0 to 500 on the X-axis of basic dance videos of lock dance (a) and ballet jazz dance (b). In (a), both the segmentation probability and the ground truth label has three peaks at approximately 110, 240, and 380 frames, which are detected as the segmentation point. In (b), the ground truth label peaks at approximately 150 and 310 frames, while the segmentation probability peaks at approximately 45, 100, 160, and 210 frames.}
    \label{fig:exampleBasic}
    \vspace{-15pt}
\end{figure}

\begin{figure}[tbp]
    \centering
    \subfigure[A success case in the advanced videos (house dance).]{
        \includegraphics[width=\hsize]{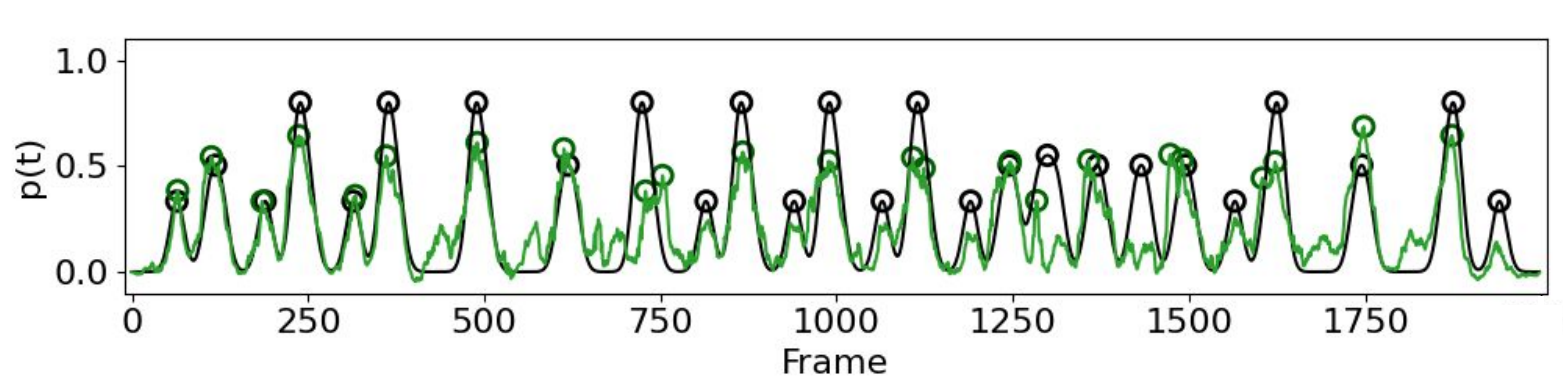}
    }
    \subfigure[A failure case in the advanced videos (street jazz dance).]{
        \includegraphics[width=\hsize]{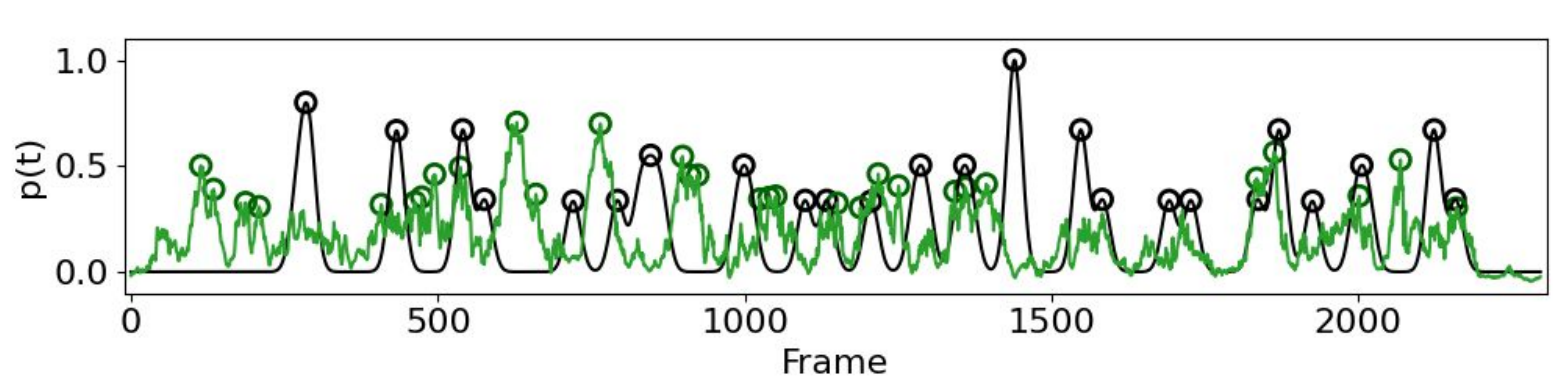}
    }
    \caption{Examples of estimating the segmentation points of the advanced videos with the proposed model.}
    \Description{Line graph showing an estimated segmentation probability and a ground truth label from 0.0 to 1.0 on the Y-axis against video frames on the X-axis of advanced dance videos of house dance (a) and street jazz dance (b). In (a), segmentation points detected from the estimated probability and the ground truth label are almost the same. On the other hand, in (b), estimated segmentation points are way off from the ground truth points.}
    \label{fig:exampleAdvanced}
\end{figure}

\autoref{tab:performanceVA} shows the performance result of the proposed model, giving the mean values and standard deviations of $D_0, \ldots, D_9$.
From the result, we confirm that the proposed model correctly estimated approximately 80\,\% of the segmentation points.
\begin{table}[tbp]
    \centering
    \caption{Mean and standard deviation of the numerical performance of the proposed model trained with each data.}
    \begin{tabular}{cccc}
        \toprule
        Precision & Recall & F-measure & Test loss \\
        \midrule
        $0.846 \pm 0.018$ & $0.754 \pm 0.020$ & $0.797 \pm 0.013$ & $0.097 \pm 0.004$ \\
        \bottomrule
    \end{tabular}
    \Description{Table showing the mean and standard deviation of the performance of the proposed model in four metrics. The precision is 0.846 ± 0.018, the recall is 0.754 ± 0.020, the F-measure is 0.797 ± 0.013, and the test loss is 0.097 ± 0.004.}
    \label{tab:performanceVA}
    \vspace{-15pt}
\end{table}
%
\autoref{fig:exampleBasic} and \autoref{fig:exampleAdvanced} show examples of the estimation result with data $D_1$, which performed the best.
In \autoref{fig:exampleBasic}a\footnote{\url{https://aistdancedb.ongaaccel.jp/v1.0.0/video/10M/gLO_sBM_c01_d15_mLO3_ch08.mp4}}, a dancer repeats a dance movement called ``skeeter rabbit'' four times, and the model can correctly segment it into four segments.
The model also correctly estimated most of the segmentation points in an advanced video\footnote{\url{https://aistdancedb.ongaaccel.jp/v1.0.0/video/10M/gHO_sFM_c01_d20_mHO1_ch09.mp4}}, as shown in \autoref{fig:exampleAdvanced}a.
On the other hand, the model failed to estimate the segmentation points in the videos of \autoref{fig:exampleBasic}b\footnote{\url{https://aistdancedb.ongaaccel.jp/v1.0.0/video/10M/gJB_sBM_c01_d08_mJB4_ch01.mp4}} and \autoref{fig:exampleAdvanced}b\footnote{\url{https://aistdancedb.ongaaccel.jp/v1.0.0/video/10M/gJS_sFM_c01_d03_mJS2_ch03.mp4}}.
In the video of \autoref{fig:exampleBasic}b, a dancer repeats a turn movement called ``chaines.''
The true segmentation points are when the dancer puts the dancer's left heel on the ground, but the proposed model segmented the video during the turn.

\subsection{Ablation Study}
\begin{figure}[tbp]
    \centering
    \includegraphics[width=0.8\hsize]{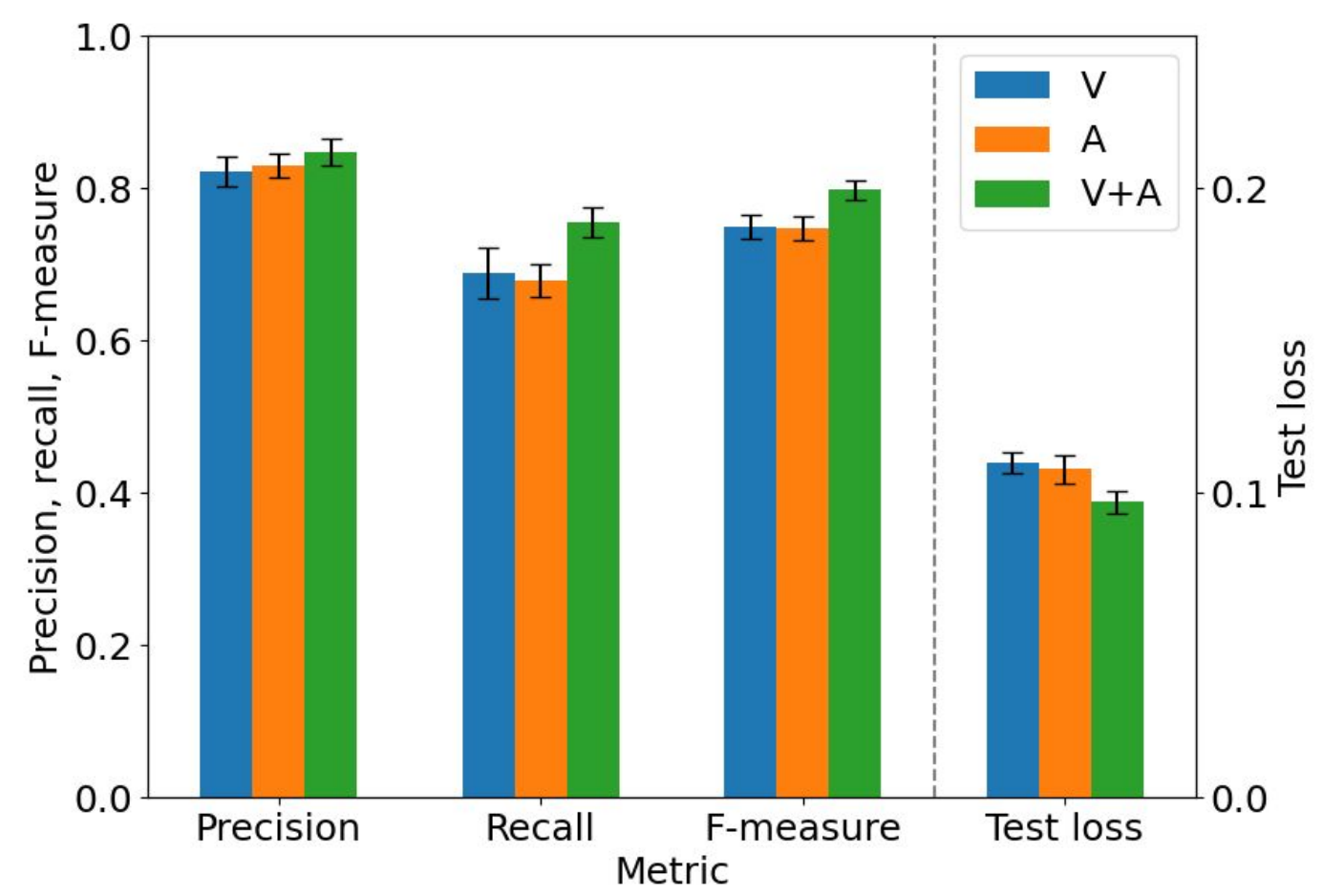}
    \caption{Mean performance of each model. Error bars show the standard deviations. V+A is the proposed model. V only uses the visual feature, and A only uses the audio one.}
    \Description{Bar graph showing the performance of each model (V, A, V+A) in four metrics (precision, recall, F-measure, test loss). The precision, recall, and the F-measure of each model are around 0.8, and the test loss is around 0.1. The precision of V+A is slightly higher than V and A. The recall and the F-measure of V+A are higher than the other two without error bar overlaps. The test loss of V+A is the lowest without error bar overlaps.}
    \label{fig:performance}
\end{figure}

We compared the performance of the proposed model with two other models.
The two models are the same as the proposed model except that one uses only the visual feature, and the other uses only the audio feature.
\autoref{fig:performance} shows the precision, recall, F-measure, and test loss of each model.
The V+A is the proposed model, which uses both visual and audio features.
The V is the model that only uses the visual feature, and the A only uses the audio feature.
We also ran $t$-tests at a 5\,\% significance level between the proposed model and the other two.
The $p$-values indicate the proposed model achieved significantly higher precision, recall, and F-measure, and lower test loss than the other two. 

\section{Application} \label{sec:app}

\begin{figure}[tbp]
    \centering
    \includegraphics[width=\hsize]{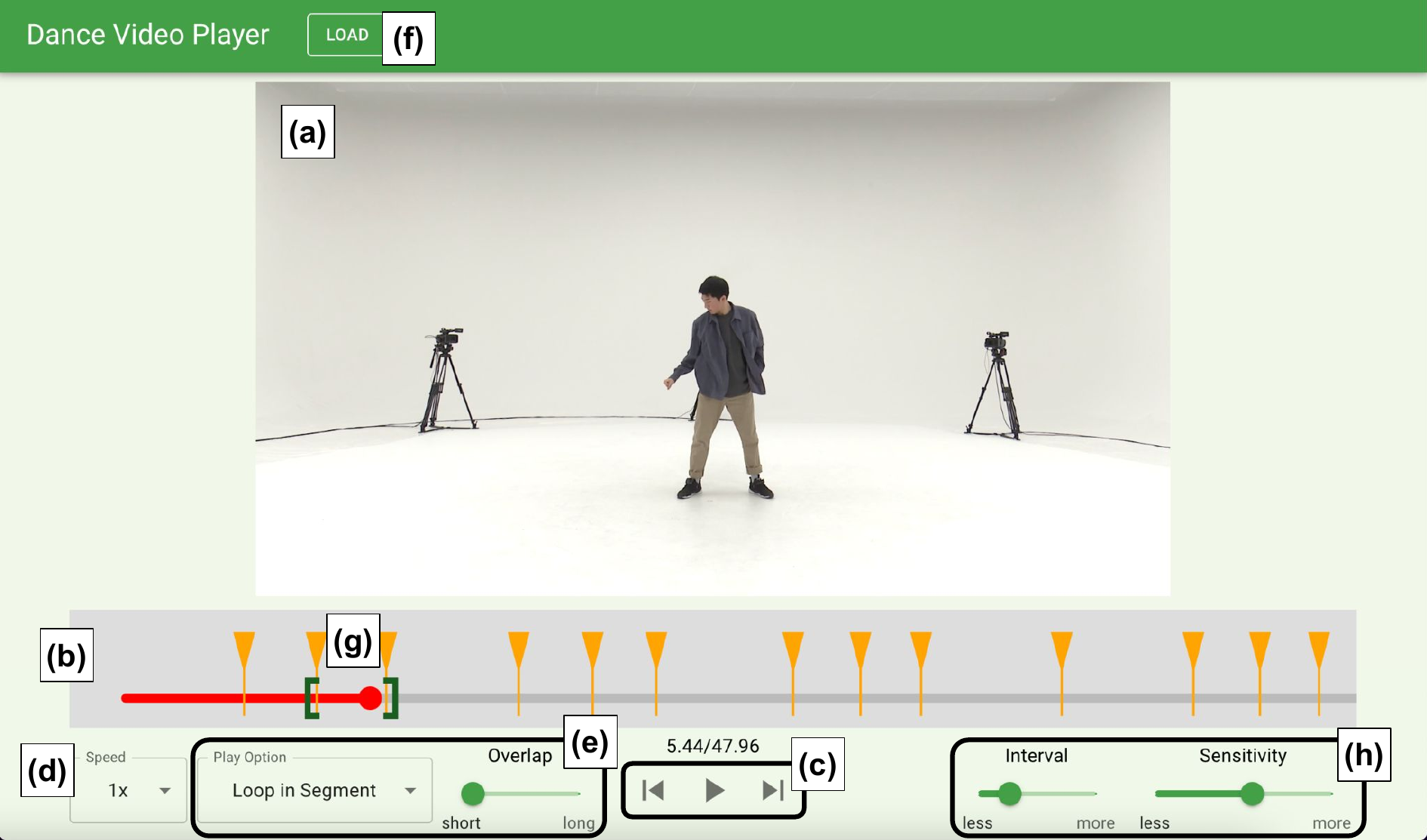}
    \caption{User interface of the application. (a) Dance video. (b) Seek bar and segmentation points. (c) Playback and skip buttons. (d) Pull-down to change the playback speed. (e) Pull-down to change the playback mode and slider to change the margin of the loop segment. (f) Button to load a video. (g) Square brackets for the loop segment. (h) Sliders to adjust the parameters for the peak picking.}
    \Description{User interface of our application. A dance video is in the center (a), and a seek bar with orange markers is below the video (b). Playback and skip buttons are on the lower center (c). A pull-down to change the playback speed is on the lower left (d). On the lower left, there are also a pull-down to change the playback mode and a slider labeled “Overlap” (e). This interface also has a header titled “Dance Video Player” with a load button (f). There are square brackets from the second to the third marker on the seek bar (g). On the lower right are sliders labeled “Interval” and “Sensitivity” (h).}
    \label{fig:app}
    \vspace{-5pt}
\end{figure}

We developed an application to help dancers practice choreography using the proposed method (see \autoref{fig:app}).
As with the annotation tool described in Section~\ref{sec:annotationUI}, this application consists of a main panel displaying a dance video (\autoref{fig:app}a), a seek bar (\autoref{fig:app}b), playback and skip buttons (\autoref{fig:app}c), a pull-down to change the playback speed (\autoref{fig:app}d), a pull-down to change the playback mode (\autoref{fig:app}e), and a button to load a video (\autoref{fig:app}f).
Orange markers on the seek bar are segmentation points estimated by the proposed method.

The application has two playback modes: {\it Nonstop} and {\it Loop}.
The {\it Nonstop} mode is the same as the annotation tool.
The {\it Loop} mode displays square brackets on the application representing the loop segment (\autoref{fig:app}g), allowing the user to easily play the segment in a loop.
The loop segment is a range between two segmentation points, and the user can choose which segment to play by changing the current playback time.
In addition, the loop segment has a margin: the application plays the segment from a few seconds before the start frame of each segment to a few seconds after the end frame of the segment.
This margin helps users understand how the movement is connected from one segment to the next.
The initial length of the margin is 0.2 seconds, but it is possible to change the length with a ``\textit{Overlap}'' slider (see \autoref{fig:app}e).

\begin{figure}[tbp]
    \centering
    \includegraphics[width=\hsize]{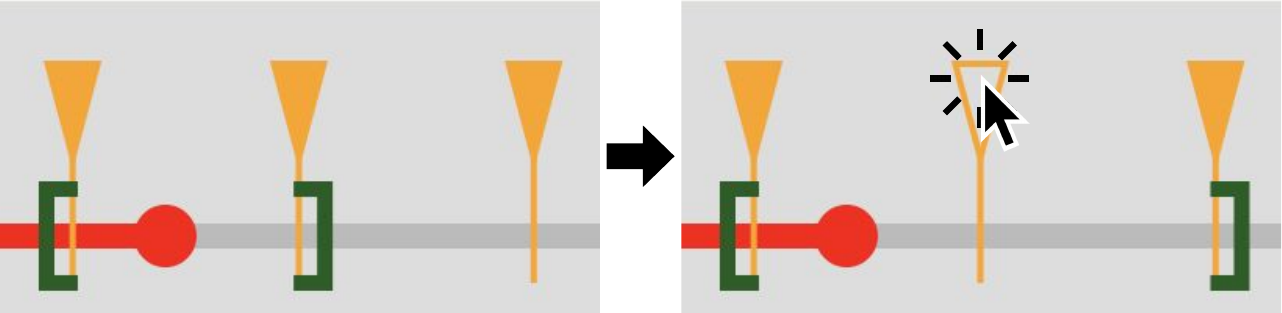}
    \caption{Deactivation of the segmentation point. The middle marker is deactivated when clicked, and the segments on either side of it are merged.}
    \Description{Screenshots showing an operation to deactivate markers. In the left figure, there are three filled markers on the seek bar and square brackets from the first to the second marker. The right figure is almost the same as the left, but the second marker turns transparent because it is clicked. Furthermore, the position of the end of the bracket changes from the second to the third marker.}
    \label{fig:merge}
    \vspace{-10pt}
\end{figure}

The user can also adjust the segmentation points using the sliders in \autoref{fig:app}h.
Increasing the value of the ``\textit{Sensitivity}'' slider lowers the peak picking threshold~$h$ described in Section~\ref{sec:peak}, allowing the application to detect more segmentation points.
In addition, the user can change the minimum distance between segmentation points with the ``\textit{Interval}'' slider.
When setting the distance value~$d$, the application automatically reduces the estimated segmentation points by greedy removing several points within~$d$ frames of the previous points.
The initial values of the threshold $h$ and the distance~$d$ are $0.25$ and $60$, respectively.
Furthermore, each segmentation point can be deactivated by clicking the marker.
The deactivated markers are not used as segmentation points, and the segments on either side of the deactivated marker are merged, as shown in \autoref{fig:merge}.

\section{Discussion}

The estimation results described in Section \ref{sec:evalResult} provide some insights about what choreography is easy to segment.
In the test data of $D_1$, the F-measure was 1.0 in many videos of house dance, LA-style hip-hop, and lock genres.
The main reason is dance movements in these genres can simply be segmented at the beat timings, and the choreography often contains basic moves in the genre.
In contrast, in the case of the test data of $D_1$, the ballet jazz and street jazz videos accounted for 60\,\% of the worst 20 on the F-measure.
From these results, it is thought that dance movements in these genres are smoothly connected, without regard to beat timings, making the segmentation difficult.

\begin{figure}[tbp]
    \centering
    \includegraphics[width=\hsize]{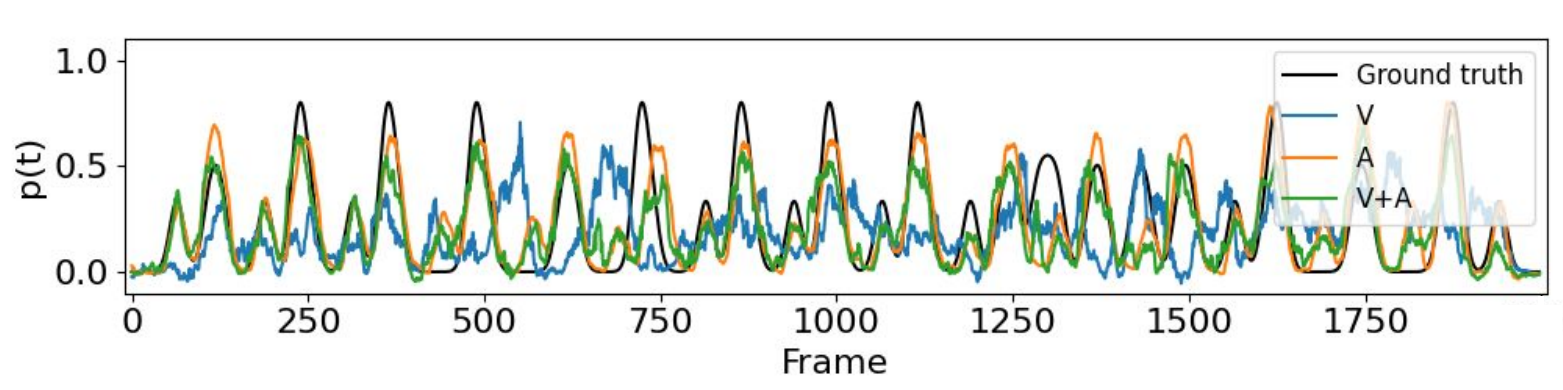}
    \caption{An example of comparing the estimated segmentation probability scores with the ground truth.}
    \Description{Line graph showing segmentation probabilities estimated by each model (V, A, V+A) and a ground truth label from 0.0 to 1.0 on the Y-axis against video frames on the X-axis of an advanced dance video. The segmentation probabilities estimated by A and V+A are similar to the ground truth label, while the probability estimated by V is not.}
    \label{fig:comparisonGraph}
    \vspace{-10pt}
\end{figure}

Next, we discuss the effectiveness of the visual and audio features.
As shown in \autoref{fig:performance}, 
the proposed model performed significantly better than the visual-only model and the audio-only model.
One possible reason is that dance choreography itself corresponds to music, but various choreographies are often designed from single music.
That is, visual information is important to deal with such differences in choreography.
In addition, we compare the segmentation probability scores based on each model trained with $D_1$ for a video\footnotemark[4].
Note that the video contains several movements that were not included in the videos used in the training step.
As shown in \autoref{fig:comparisonGraph}, we confirm that the visual-only model cannot correctly estimate the probabilities, but the proposed model and the audio-only model were able to estimate probability scores close to the ground truth.
In summary, combining visual and audio features is important for the segmentation task.

\section{Conclusion}

The paper has proposed a method for automatically segmenting dance videos into short movements.
Our method first extracts visual and audio features from an input dance video.
The visual feature is computed by passing the bone vectors of the dancer in the video to a fully connected network, and the audio feature is obtained by convolving the Mel spectrogram of the music in the video.
These features are then passed to a TCN~\cite{Bai2018TCN}, and the network outputs a segmentation probability score in each frame.
We constructed a dataset by manually annotating segmentation points for the videos in the AIST Dance DB~\cite{Tsuchida2019AISTDanceDB}.
We define the ground truth by representing the segmentation points of each annotator as the sum of Gaussian distributions, named segmentation probability.
The evaluation result using this dataset shows that the proposed model can correctly segment the street dance videos, in which the many movements are aligned to beats.
In addition, the result of the ablation study suggests the effectiveness of both the visual and audio features.
Furthermore, we developed an application using our segmentation method for supporting video-based practice.
This application enables users to adjust the segmentation points according to the users' preferences and watch each segment on a loop, so users can efficiently practice the choreography.



\begin{acks}
This work was supported by JST, CREST Grant Number JPMJCR17A1, and JSPS Grant-in-Aid 23K17022, Japan.
\end{acks}

\bibliographystyle{ACM-Reference-Format}
\bibliography{myref}


\end{document}